\documentclass[11pt]{article}
\usepackage{acl2016}
\usepackage{times}
\usepackage{latexsym}
\usepackage{amsmath}
\usepackage{graphicx}
\usepackage[english]{babel}
\usepackage[utf8x]{inputenc}
\usepackage[T1]{fontenc}

\aclfinalcopy 

\title{A Dual Encoder Sequence to Sequence Model for Open-Domain Dialogue Modeling}

\author{Sharath T. S.\\
	    UC Santa Cruz\\
	    CA, USA\\
	    {\tt sturuvek@ucsc.edu}
	  \And
	Shubhangi Tandon\\
  	UC Santa Cruz\\
	CA, USA\\
	{\tt shtandon@ucsc.edu}
     \And
	Ryan Bauer\\
  	UC Santa Cruz\\
	CA, USA\\
	{\tt rbauer@ucsc.edu}}

\date{}

\begin{document}

\maketitle

\begin{abstract}
Ever since the successful application of sequence to sequence learning for neural machine translation systems \cite{sutskever2014sequence}, interest has surged in its applicability towards language generation in other problem domains. Recent work has investigated the use of these neural architectures towards modeling open-domain conversational dialogue, where it has been found that although these models are capable of learning a good distributional language model, dialogue coherence is still of concern. Unlike translation, conversation is much more a one-to-many mapping from utterance to a response, and it is even more pressing that the model be aware of the preceding flow of conversation. In this paper we propose to tackle this problem by introducing previous conversational context in terms of latent representations of dialogue acts over time. We inject the latent context representations into a sequence to sequence neural network  in the form of dialog acts using a second encoder to enhance the quality and the coherence of the conversations generated. The main task of this research work is to show that adding latent variables that capture discourse relations does indeed result in more coherent responses when compared to conventional sequence to sequence models.

\end{abstract}
\section{Introduction}
Our task is to develop an enhancement for a sequence to sequence generative model for dialogue that provides for globally coherent responses. We propose to do this by making use of a history of Dialogue Acts to obtain a global context for the model. The task of open-domain, casual conversation with a chat bot requires a model of dialogue coherency such that a provided system-response is coherent in both the immediate context as well as the broader context of the entire conversation so far. Current models for dialogue coherence, for both retrieval-based and generative systems in the open-domain setting, do not focus on incorporating more of the prior conversation into the process of generating the current system-response. For neural dialogue models, the idea of incorporating prior context during generation of a response started off with \cite{sordoni2015neural}. Generative neural models, in particular, have seen a recent surge in interest towards incorporating aspects of global conversation into current sequence to sequence models, which historically only handled generating a response to a single given utterance. The methods by which to encode these “global cues” currently varies in terms of topic, sentiment, dialogue act, etc., and as such the task of developing such augmentations for these networks and determining which makes for more coherent conversations is still under exploration. 

Additionally, the difficulty of developing models for conversation is compounded by the lack of established metrics for evaluation of performance. Established metrics such as BLEU and METEOR, which measure various types of alignment between a system output and a target phrase, make intuitive sense in the domains of translation and summarization, but breakdown in the setting of conversation, where many generated responses could serve as a quality response. Measures like perplexity that measure the predictive fit of a model on some held out set of utterances do not necessarily correlate well with human ideas about what constitutes a quality conversation. Consequently, recent work has investigated new ideas for evaluating dialogue models, such as information gain across turns or discriminative adversarial evaluation, but the landscape is rapidly shifting. This paper does not investigate the evaluation problem in depth, but the issue has been kept in mind.

\section{Related Work}
The task of open domain conversation has gradually evolved from a simple retrieval based method based using similarity metrics to a sequence to sequence generation problem \cite{serban2016building} \cite{vinyals2015neural} and also an ensemble of both \cite{song2016two}. Retrieval based methods, while effective are limited in their capacity of only producing responses that it has seen before. A simple advantage is that these methods do not produce output that is syntactically incorrect. A generative model on the other hand learns a language model, which is a distribution on words in the vocabulary conditioned on the previous ones and is capable of producing sentences not seen in the training set.
Though a generative model is capable of generating new output, it tends to generate short and most generic of responses. 
This is the one of the main reasons why the state of the art for machine translation cannot be directly ported to the task of coherent dialogue generation. A particular utterance can map to multiple coherent responses and the conventional loss functions such as cross entropy aren't well suited for this task. Such loss functions make the model expect a particular response penalize any others even though they are coherent in terms of semantics and context.

There has been a lot of research currently to address these loss functions, where researchers have tried incorporating reinforcement policies where the model is rewarded for producing more diverse and longer responses \cite{li2016deep}. Adversarial strategies have also been adopted where a discriminative network is stacked on a generator and the discriminator classifies whether the generated response resembles a human response or not \cite{li2017adversarial}.

With all this said, there has been very little focus on incorporating the context of the conversation that has happened so far in order to generate responses. \cite{sordoni2015neural} propose a naive approach where they incorporate the previous set of utterance, responses as a bag of words model and use a feed forward neural network to inject a fixed sized context vector as a into the LSTM cell of the encoder. \cite{ghosh2016contextual} proposed a modified LSTM cell with an additional gate that incorporates the previous context as input during encoding. The weights of the gate are learned exactly in the same way the weights for the input, forget and output gates are learned.

\cite{xiong2016neural} propose a context/topic sensitive question answering system where a convolution neural network is pre-trained to predict one among 40 topics. The convolutional model is given as input a window of the previous conversation and the hidden state of the encoder before the softmax layer is fed as input(context) to a sequence to sequence model along with the current utterance for coherent answer generation.

We propose to adopt the above idea for open domain language generation, and as context, perform dialogue act classification and feed the hidden state to the sequence to sequence model to generate coherent responses.

\section{Proposed Approach}
\subsection{Baseline Models}
\subsubsection{Baseline 1}
For our primary baseline, we utilize a Bidirectional Sequence to Sequence Model with Attention,
an encoder-decoder based sequence to sequence neural model for generating a system response given an utterance. Both the encoder and decoder are LSTM based Recurrent Neural Networks. The encoder is bidirectional and the decoder has a multiplicative (Luong) attention mechanism. Input utterances are padded, binned and at inference time the response is dynamically unrolled. This particular baseline model is the foundation of the current state of the art in neural machine translation, very similar to the architecture described in \cite{bahdanau2014neural}
\subsubsection{Baseline 2}
The second Baseline is a replica of Baseline 1 in architecture, but the the input to the encoder of the model is modified to be a window of previous utterances concatenated together.  
\subsection{Context Encoder}\label{contextencoder}
\begin{figure*}
\centering
\includegraphics[width=\linewidth, height=8cm,keepaspectratio]{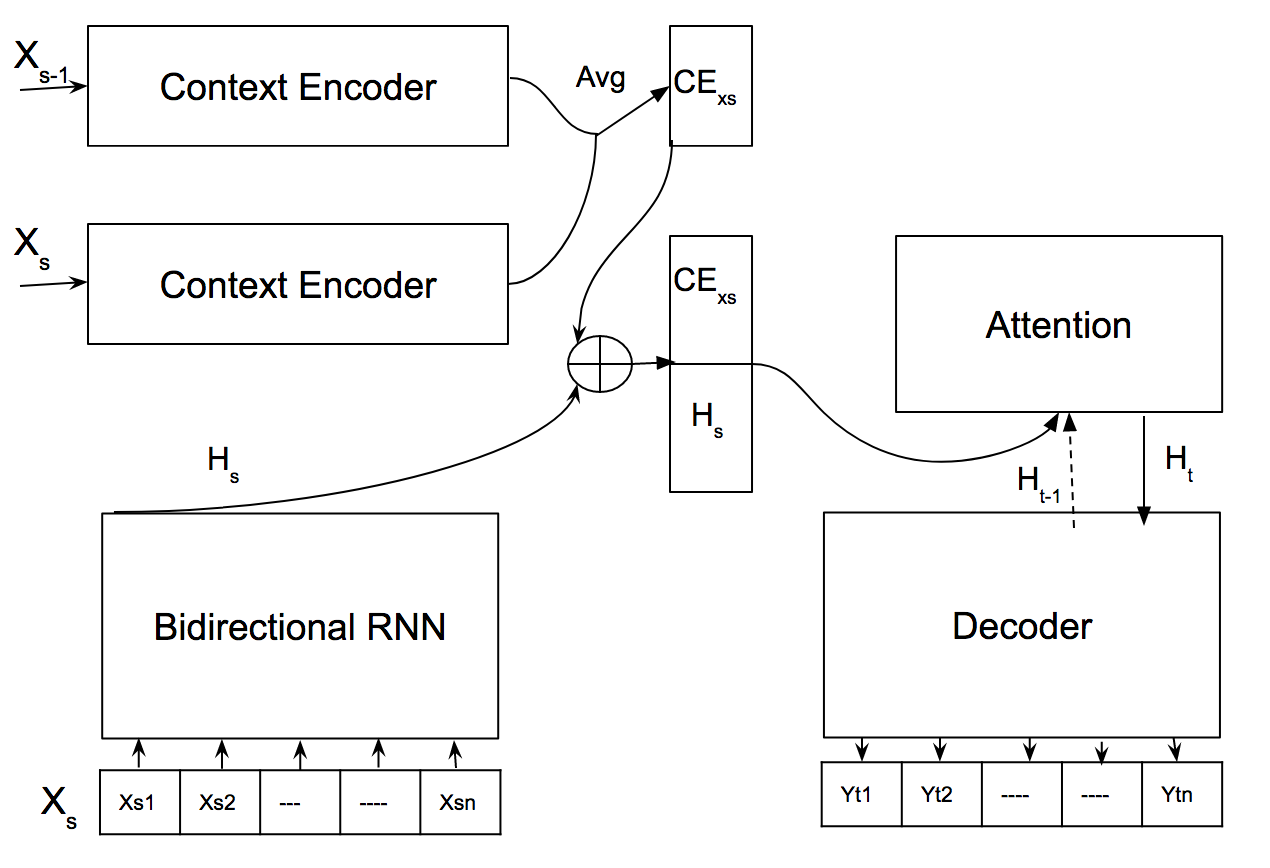}
\caption{\label{fig:model}Context Seq2Seq Model
}
\end{figure*} 
It is evident that both of our baselines do not take the context, i.e., any of the previous conversation that has happened so far into picture when generating a response the current utterance. To incorporate for this, we propose to have a Convolutional neural network which is pre-trained to predict a dialogue act given an input utterance \textbf{$X_{s}$} as in \ref{fig:model}. The context model encodes the input sentence in a space that can easily discriminate it among different dialogue act classes. The hidden layer of the context model, just before the softmax classification layer is fed as context to the sequence to sequence model to generate coherent responses.
\subsubsection{Architecture}
The context model consists of an embedding layer of dimension 128 with a maximum allowed sequence length of 25. We perform convolutions on entire words(embedding size of 128) of window length 3, 4, 5, 6 and 8, with 128 different filters and a stride of 1 word. The output of the convolutions are max-pooled and concatenated together. It then consists of a feed forward layer of size 512, which would later be extracted and fed into the sequence to sequence model as global context. It then consists of a softmax layer which classifies as one among 10 discourse tags. The performance of the model in terms of it's confusion matrix is shown in \ref{fig:confusion_matrix}. It also achieves an accuracy of 71.6

\subsection{Context Seq2Seq}
In our extended model \ref{fig:model}, we introduce another component on the input side of the model which we refer to as the “context encoder” as described in \ref{contextencoder}. This component is responsible for generating a representation of the conversation context \textbf{$CE_{XS}$} to the current utterance and making it available to the decoder. We represent context as a truncated history of dialogue acts, and propose the below algorithm:
\begin{enumerate}
\item Train the context model to predict the dialogue act given an utterance.
\item Extract the hidden layer(prior to softmax) of the context model as a vector to provide to the decoder model as context. We generate this vector for the preceding two dialogue pairs, and the final context vector is an average of these vectors, denoted by \textbf{$H_{S}$} in \ref{fig:model}.
\item The resulting context vector is concatenated with every encoder hidden state denoted by \textbf{$CE_{XS}$} in \ref{fig:model} making for attention keys of dimension 1024, since the encoder hidden state and the context vector are of size 512 each.
This vector is then passed to a Feed-Forward network that reduces it to a vector of size 256 which is the hidden size of the decoder denoted by \textbf{$H_{1}$} ... \textbf{$H_{t}$} in \ref{fig:model}.
\item Teacher forcing is used during training and we implement beam decoding at inference time to get a response to an utterance that is conditioned on the prior conversation. During inference, we perform both greedy decoding and consider beams of size 3, with inclination towards picking sequences that are least probable.  

\end{enumerate}

\section{Data}
We have employed the following datasets to train our models:
\begin{enumerate}
\item Switchboard Corpus for Context Encoder. 
We use the Switchboard Dialogue Act Corpus(SwDA) that extends the Switchboard-1 Telephone Speech Corpus and has turn/utterance level dialog act tags. The corpus entails around 2400 two sided telephonic conversations that range over 70 topics and result in over 200,000 response-utterance pairs . 

We use the SwitchBoard Corpus for training
our Context Encoder to predict a Dialog Act tag for for a given utterance. 
These Dialogue Act Tags are an extension of the dialogue acts described in \cite{stolcke2000dialogue} which represent shallow discourse structure of adjacent conversation pairs. The tags summarize higher level action associated with the utterance based on its syntactic, pragmatic and semantic information. Examples of dialogue acts include asking a question or accepting a prior statement. More examples of dialogue act categories along with their counts in the dataset have been provided in Table 1. Note that, we have condensed the existing 43 discourse tags to a set of 10 general discourse acts.

\begin{table*}
\begin{tabular}{|c|c|c|}
\hline
\textbf{Dialog Act Class} & \textbf{Count} & \textbf{Example}\\\hline
Accept & 21149 & That is exactly it!
\\\hline
Non-Opinionated & 74992 & Me, I'm in the legal department and I have been working on it since.\\\hline
Backchannels & 71235 & Uh-huh.\\\hline
Opinionated & 26770 & I think that is a great idea and we should try it out.\\\hline
Questions & 10955 & How are you feeling today?\\\hline
Summarize(Repeat) & 2305 & Oh, you mean you switched schools for the kids.\\\hline
Reject(Oppose) & 1379 & Well, I don't think so.\\\hline
Conventional response & 3710 & Well, It was nice talking to you.\\\hline
Non verbal & 1891 & [Laughter]\\\hline
Other & 2148 & Well give me a break, you know.\\\hline
\end{tabular}
\caption{Dialog Act Data and Classes in Switchboard Corpus}
\end{table*}

\begin{figure}
\centering
\includegraphics[width=\linewidth, height=11cm,keepaspectratio]{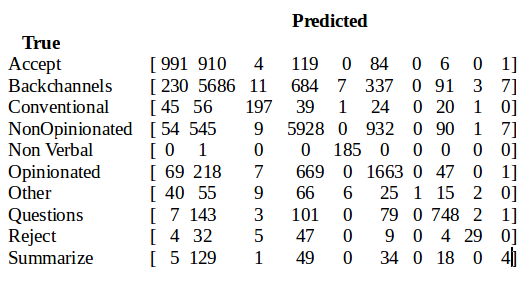}
\caption{\label{fig:confusion_matrix}Performance of CNN on validation set
}
\end{figure}

\item Cornell Film Corpus for Dialogue Model. 
We use the Cornell Film corpus to train our baseline dialogue models as well as our ContextSeq2Seq model. The Cornell Film corpus consists of a collection  of movie scripts, from which we extract conversations between characters from which our models can learn. Movie dialogue is a reasonable approximation of open-domain conversation, and the fact that we can extract speaker-aligned utterances from the corpus makes it attractive for training dialogue models. Once we have extracted utterances and responses, we end up with ~240,000 examples in our training set for all models.

We note that we also performed some training on dialogue from the OpenSubtitles2016 dataset, but those results have not been included here as the significantly larger size of the dataset makes training very expensive. Additionally, this data is not speaker-aligned, so we employed the very rough heuristic of splitting by sentence to form utterance response pairs, as done by Vinyals and Le[7].
\end{enumerate}

\begin{figure*}
\centering
\includegraphics[width=\linewidth, height=\textheight]{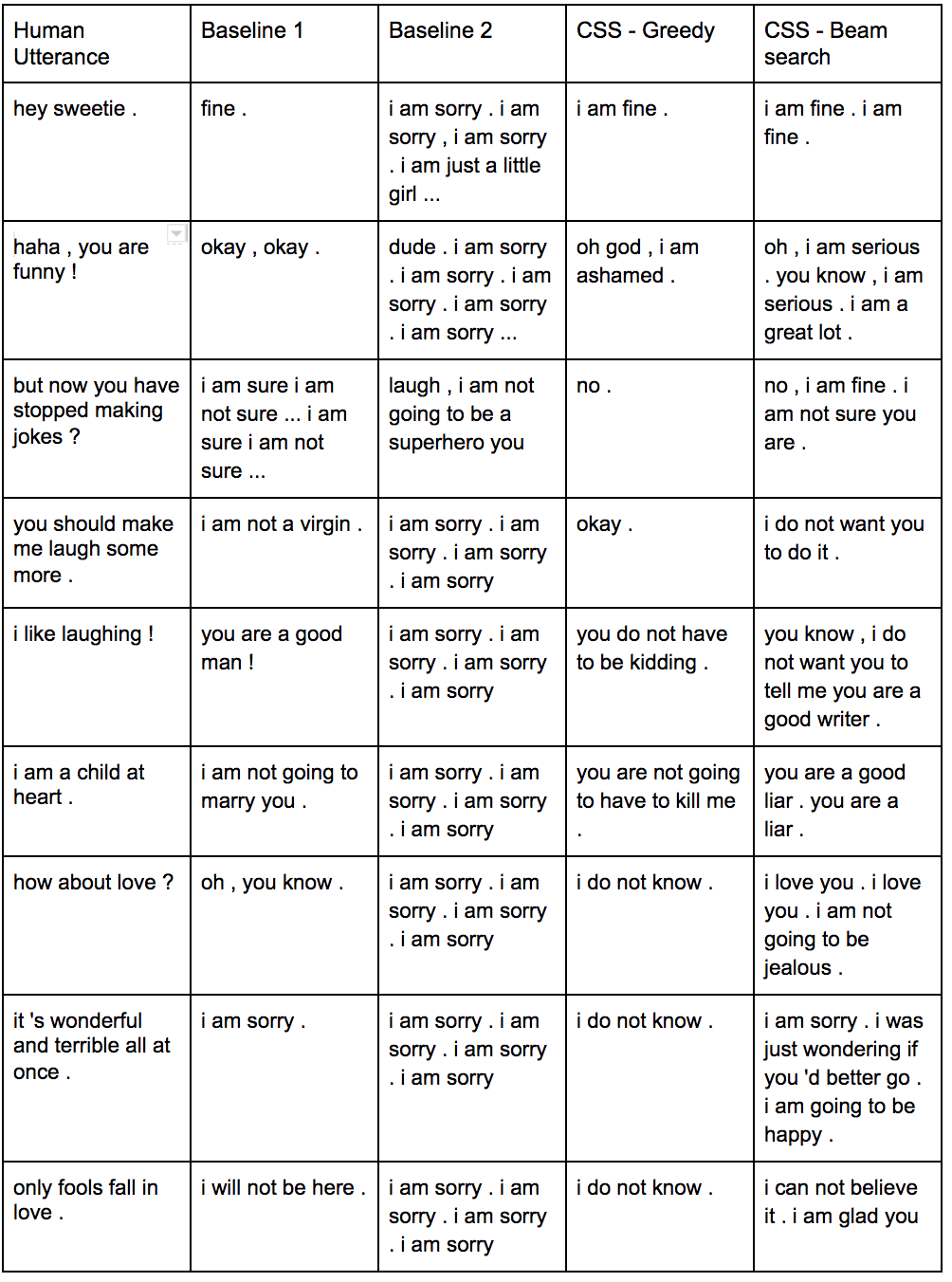}
\caption{\label{fig:Conversations}Reponses generated by our models and Human Utterances Sampled from Tick-Tock
}
\end{figure*}

\section{Empirical Results}
It is clearly evident that from the \ref{fig:Conversations} that our model outperforms the two baselines in terms of diversity of responses and coherency. This asserts that fact that a dialogue is not simply a one to one mapping of utterance response pairs, and including prior context in terms of dialogue acts helps the model learn more coherent and diverse responses.

\section{Observations}

\begin{figure}
\centering
\includegraphics[width=\linewidth, height=7cm,keepaspectratio]{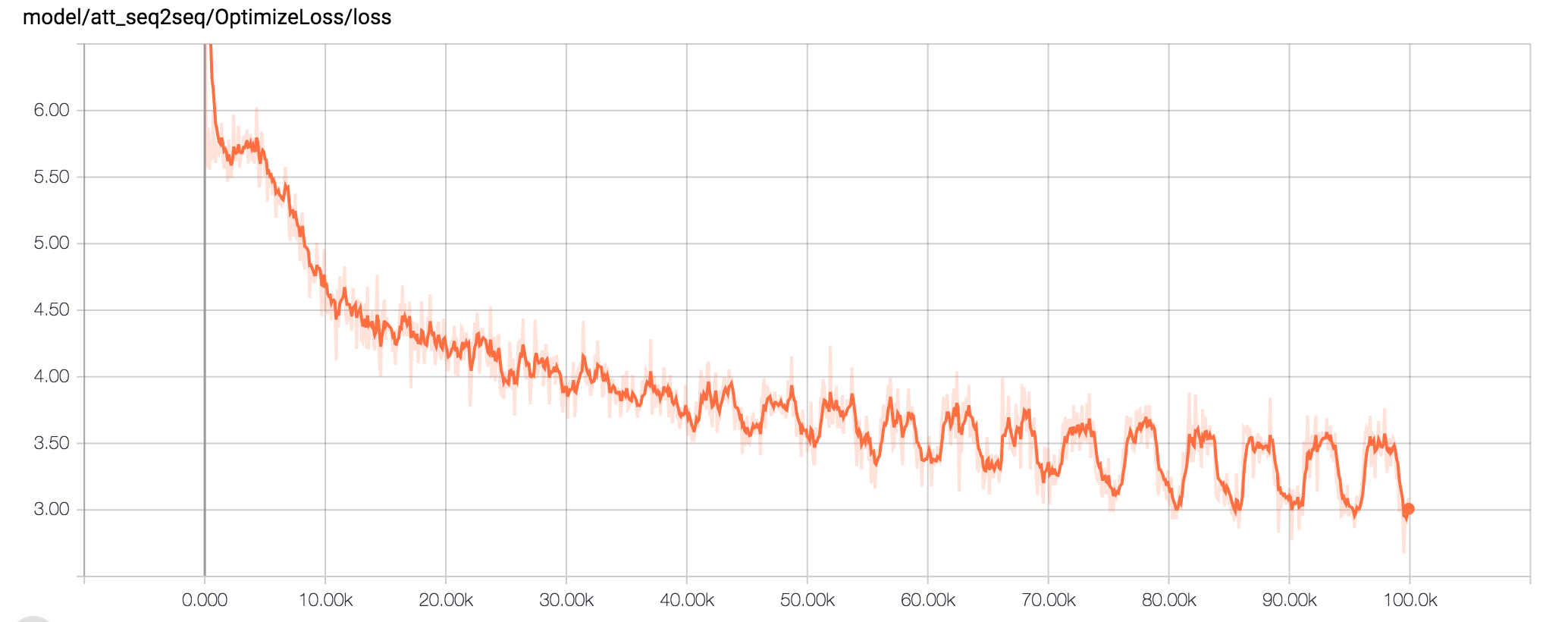}
\caption{\label{fig:loss_b1}Training Loss Curve for Baseline-1 model
}
\end{figure}

\begin{figure}
\centering
\includegraphics[width=\linewidth, height=7cm,keepaspectratio]{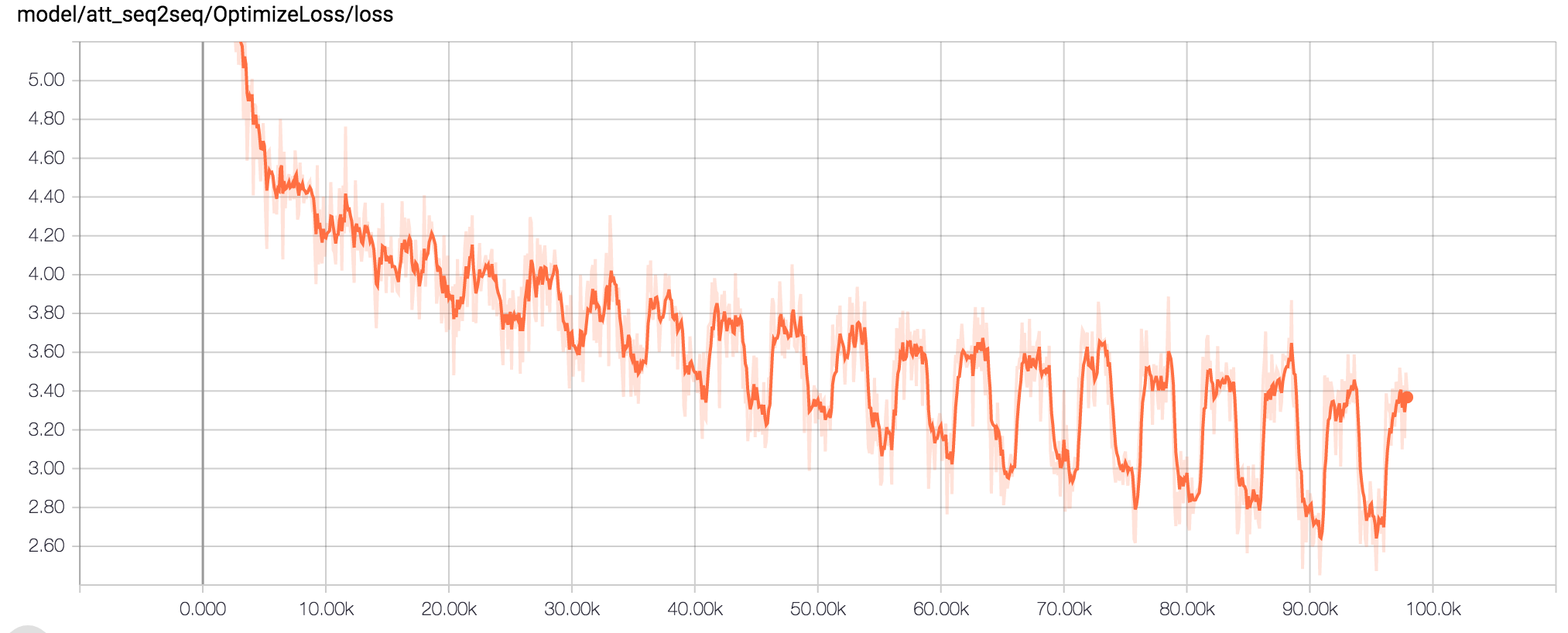}
\caption{\label{fig:loss_b2}Training Loss Curve for Baseline-2 model
}
\end{figure}

\begin{figure}
\centering
\includegraphics[width=\linewidth, height=7cm,keepaspectratio]{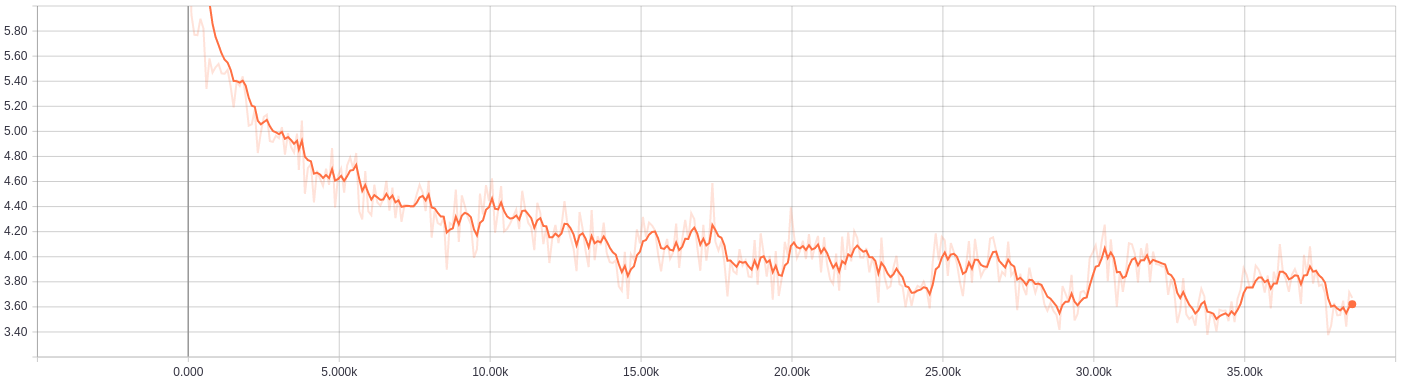}
\caption{\label{fig:loss_train_cs}Training Loss Curve for Context Seq2Seq
}
\end{figure}

\begin{figure}
\centering
\includegraphics[width=\linewidth, height=7cm,keepaspectratio]{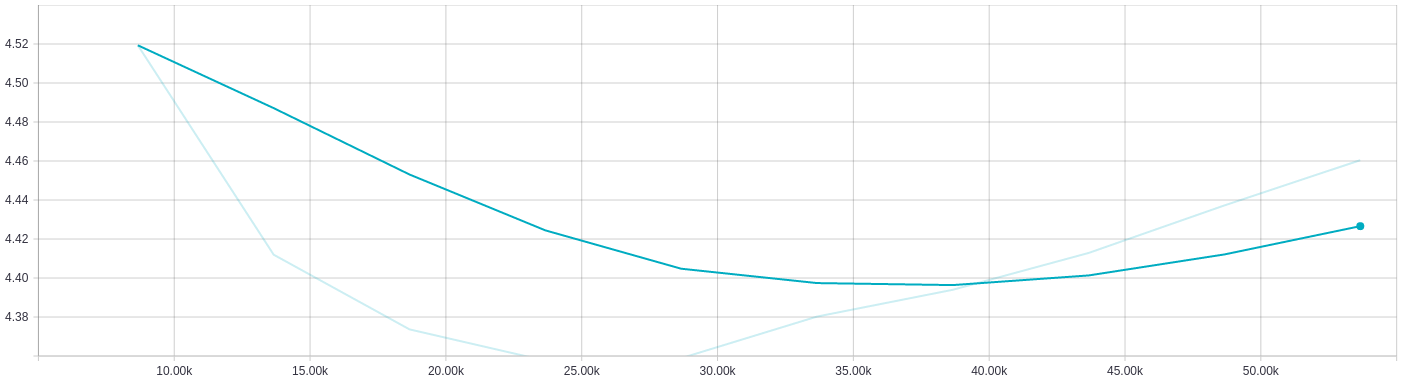}
\caption{\label{fig:loss_val_cs}Validation Loss Curve for Context Seq2Seq
}
\end{figure}

When training all three models on the Cornell movie corpus, the loss for gradually decreases as in \ref{fig:loss_b1} and \ref{fig:loss_b2} \ref{fig:loss_train_cs}, but the validation losses increase after a certain point \ref{fig:loss_val_cs}. Evaluating the model at the lowest point on the loss curve doesn't give us an better results. It clearly supports the claim that cross entropy loss if not well suited for the problem of dialogue generation.

The baseline1 gives responses that are syntactically correct, but not coherent to the utterance in any way. It in facts tends to generate very generic responses as it has no idea of the context of the conversation.

The baseline2 perform worse and proves the fact that simply concatenating a window of the previous conversations together in fact makes the model worse. It is a know fact that thought LSTMs can theoretically remember over a long sequence, it is not true in practice.

The Context Seq2Seq models significantly outperforms both the baselines in the sense that it's responses are not generic and are more coherent with respect to the context of the conversation. The results are presented in \ref{fig:Conversations}.

Another observation is that the responses tend to get shorter with more epochs on the training set, as in \ref{table:automated}. This can be attributed to the small size of the Cornell Corpus.

\begin{table*}
\begin{tabular} {|c|c|c|c|c|}
\hline
\textbf{Model}&\textbf{Median Length }&\textbf{Mean Length }&\textbf{Diversity } &\textbf{Specificity (Mean)}\\\hline
Baseline1 &6 &9.11 &0.007 &0.055002  \\\hline
Baseline2 &19.5 &49.73 &0.0013 &0.454314\\\hline
Context Seq2Seq (CSS) &11.0 &17.5 &0.0038 &0.027218\\\hline
Context Seq2Seq With Beam (CSS-B) &4.0 &4.75 &0.0135 &0.080632\\\hline
\end{tabular}
\caption{\label{table:automated} The Average Length and Specificity score for responses generated by the Baseline Models and the Context Model }
\end{table*}

\section{Evaluation}
For evaluation,we are comparing four models :  Baseline1 , Baseline2 and Context Seq2Seq with greedy decoding and Context Seq2Seq with Beam Decoding ( beam width : 3, length penalty: 0, chosen beam :3 ).

We trained our models with identical hyper-parameters and then evaluated them against  the human-bot conversations .We evaluate all of our models on a set of 10 conversations of 13 utterances each pulled from the Tick-Tock corpus, which simulate human to bot conversations. We remove the bot responses from the conversations ,feed the user utterances to our model and fill in with the responses generated by our models. A comparative analysis has been shown in table \ref{fig:Conversations}. We have performed Quantitative Evaluation and Qualitative Analysis on the generated output . 

\subsection{Experiments}

We performed several experiments on the decoder in terms of greedy decoding and beam decoding. We experimented with varying beam widths from 3 - 100 and length penalties of 0.2 - 2. It was observed that greater the beam width, a optimal subtree was found with resulted is a more generic response. Choosing one among the least probable beams contributed to diversity of responses. Penalizing the output length didn't affect much since we choose one among the less probable beams. Ultimately, the best hyper parameters for our model was a beam width of 3.

\subsection{Quantitative Evaluation}
For the scope of our project , standard NMT metrics such as ROUGE and BLEU are not ideal for the task of Language Generation for casual conversational tasks . ( \cite{liu2016not} 
For quantitative evaluation , we look at the mean and median lengths of generated responses, as well as their diversity and mean specificity score . 
The results for these metrics are listed in table \ref{table:automated} .
\begin{enumerate}
\item Mean Length of Response 
The length of the response is an indicator of the average length of the generated output. While a longer length doesn't necessarily guarantee that the quality of conversation is good, shorter (two words or less) lengths are an indicator of the model giving objective answers or defaulting to back channels.

We see that the mean length of Baseline2 is the highest , followed by CSS , Baseline 1 and CSS-B . The length of Baseline2 is almost 50 which is the maximum length of input utterances for our models, which indicates presence of outliers.
\item Median Length of Response 
While performing qualitative analysis of the model outputs, we noticed that the some of the responses were stuck in a loop , repeating the same phrase again and again . To account for these outliers we also calculated the median length of the generated responses. 

As with median length , we find that the median length of Baseline2 is the highest , followed by CSS , Baseline 1 and CSS-B . However, the median length of Baseline2 is now comparable to the length of CSS , confirming the presence of such outliers in Baseline 2.  
The mean and median lengths of responses generated by CSS-B are very close to each other , confirming the absence of such outliers in the this model's responses.

\item Diversity of Response 
We have measured the diversity of generated response as the number of unique unigrams generated divided by the total number of tokens (unigrams) generated by a model . This is a good metric to measure that the model is actually giving distinct responses and not repeating the same answers again and again.
								
\textit{ Diversity =   Number of Unique Tokens / Total Number of Generated Tokens}

CSS-B model has the highest diversity , followed by Baseline 1 , CSS and Baseline2 . This reinforces the assumption that Baseline2 is producing low quality outputs with very less diversity . 
\item Specificity of the generated response 
Another measure  of the quality of the generated response is to check how specific ( or \ not vague ) it is. We use Speciteller \cite{li2015specificity} , a tool designed by UPenn that predicts the sentence specificity and assigns a score between 0 and 1, 0 being extremely vague and 1 being very specific.  

We find that Baseline2 model has a very high specificity score , followed by CSS-B, Baseline 1 and CSS-B.  On further inspection we see that Baseline2 consists of responses like 'I am Sorry . I am Sorry ......' repeated several times to create long responses which Speciteller assigns a near perfect score. Since Speciteller doesn't seem to take the repetition of phrases in a sentence into account , it is not a very good metric for the evaluation of our models .
\end{enumerate}

\subsection{Qualitative Analysis}
By looking at the quantitative Analysis , we can see that Baseline2 doesn't  perform as well as the other models , with the least diversity and long-repeated sentences. In-fact the responses from Baseline1 are more coherent and grammatically consistent than Baseline2. While this would seem counter-intuitive at first because we're providing the previous utterance along with the current utterance to the model as an input , on further analysis we find that simply adding more text doesn't improve the quality of a model . 
By introducing context in the form of discourse tag , we see much better results. 
CSS-B is the best performing model , with most coherent responses followed by CSS, Baseline 1 and Baseline2. Not only are responses from CSS-B more coherent , but more engaging, unique and interesting as well . 

\section{Future Work}
Now that we have successfully incorporated context in the form of dialogue acts, further contextual features that capture the topic, sentiment of the conversation could be added to enhance the model. An evolving global topic vector like in \cite{li2016neural} would enhance the coherency with respect to topic that is carried forward in a conversation. Also, training on a larger dataset such as the OpenSubtitles would help the model learn a good language model and generalize better for coherent responses. We are also cognizant of the fact that an MLE loss function is not suited for a Dialog Generation task , and would like to experiment with Reinforcement Learning , Adverserial Networks and Variational Auto Encoders. 
Automated evaluation metrics that capture the mutual information between the utterance and the response could be explored.

\section*{Acknowledgments}
We would like to acknowledge the contribution of our guide Marilyn Walker and Teaching Assistant Shereen Oraby . 
We have also employed the code developed by Google Brain members for \cite{britz2017massive} and subsequently open-sourced, which provided a good jumping-off point for us to learn and develop enhanced seq2seq models. 

\bibliography{acl2016}
\bibliographystyle{acl2016}

\end{document}